\pdfoutput=1

\documentclass[11pt]{article}

\usepackage[preprint]{acl}

\usepackage{times}
\usepackage{latexsym}
\usepackage{amsmath}
\usepackage{multirow}
\usepackage{pifont}
\usepackage{subfig}

\usepackage[T1]{fontenc}

\usepackage[utf8]{inputenc}

\usepackage{microtype}

\usepackage{inconsolata}

\usepackage{graphicx}

\title{Rumor Detection by Multi-task Suffix Learning based on \\Time-series Dual Sentiments}

\author{%
  Zhiwei Liu\textsuperscript{1}\quad
  Kailai Yang\textsuperscript{1}{\thanks{Corresponding Author}}\quad 
  \textbf{Eduard Hovy}\textsuperscript{\textbf{2},\textbf{3}}\quad
  \textbf{Sophia Ananiadou}\textsuperscript{\textbf{1}} \\ 
    \textsuperscript{1} The University of Manchester \quad 
    \textsuperscript{2} The University of Melbourne \quad 
    \textsuperscript{3} Carnegie Mellon University \quad \\
\texttt{\{zhiwei.liu,kailai.yang,sophia.ananiadou\}@manchester.ac.uk} \\
\texttt{eduard.hovy@unimelb.edu.au}
}

\begin{document}
\maketitle
\begin{abstract}
The widespread dissemination of rumors on social media has a significant impact on people's lives, potentially leading to public panic and fear. Rumors often evoke specific sentiments, resonating with readers and prompting sharing. To effectively detect and track rumors, it is essential to observe the fine-grained sentiments of both source and response message pairs as the rumor evolves over time. However, current rumor detection methods fail to account for this aspect. In this paper, we propose MSuf, the first multi-task suffix learning framework for rumor detection and tracking using time series dual (coupled) sentiments. MSuf includes three modules: (1) an LLM to extract sentiment intensity features and sort them chronologically; (2) a module that fuses the sorted sentiment features with their source text word embeddings to obtain an aligned embedding; (3) two hard prompts are combined with the aligned vector to perform rumor detection and sentiment analysis using one frozen LLM. MSuf effectively enhances the performance of LLMs for rumor detection with only minimal parameter fine-tuning. Evaluating MSuf on four rumor detection benchmarks, we find significant improvements compared to other emotion-based methods. 
\end{abstract}

\section{Introduction}

Rumors, one type of misinformation, are typically widely circulated stories or statements that have not been confirmed or verified as fact \cite{difonzo2007rumor}. All rumors start as uncertain and are later classified as true (verified), false (debunked), or unverified (keep uncertain) \cite{zubiaga2016analysing}. The characteristics of social media that enable widespread dissemination have become the primary carriers for the propagation of rumors \cite{reshi2019rumor}. Widespread rumors can cause serious consequences in real life, including political events, the economy, and social stability \cite{cao2018automatic}. 
Figure \ref{fig:rumorsamples} shows several examples. 
As such, the need for high-performance methods to automatically detect rumors is becoming increasingly urgent.

Misinformation often triggers specific emotions/sentiments to encourage re-shares, highlighting the critical role of sentiment and emotion in misinformation detection tasks \cite{liu2024emotion}. Posts on social media typically consist of a source and corresponding comments, both of which convey sentiments. The pairing of source-response emotions/sentiments is commonly referred to as ``dual emotion'' \cite{zhang2021mining}. Notably, the sentiments expressed in comments typically change dynamically over time \cite{nguyen2012predicting}. Moreover, sentiment intensity is a fine-grained dimension of sentiment analysis, typically scored between 0 and 1, where 0 indicates strong negativity and 1 indicates strong positivity \cite{wang2018using}, which can help capture the sentiments that change dynamically over time \cite{hutto2014vader}.

Numerous studies have explored rumor detection, including time-based methods \cite{guo2022temporal,qu2024temporal,cavalcante2024early} and LLMs-based methods \cite{augenstein2024factuality,hu2024bad,wang2024llm}. Some researchers apply multi-task structures to integrate emotion information \cite{ekbal2024multitasking,jiang2024makes,kumari2024emotion} and several studies enhance LLMs through fine-tuning or RAG based on sentiment features \cite{Liu2024ConspEmoLLMCT,liu2024raemollm}. Unfortunately, most of these studies focus only on the coarse-grained sentiment of the source, ignoring fine-grained pair-wise sentiments in both source and responses. Although certain methods do address dual emotion for rumor detection \cite{luvembe2023dual,zhang2021mining,wang2024multimodal}, they overlook the dynamic changes and rely on complex structures that consume time and resources. To bridge these gaps, it is crucial to develop a framework that not only captures dual sentiment dynamics over time but also efficiently integrates them into the rumor detection process.

Prompt-tuning \cite{lester2021power} and prefix-tuning \cite{li2021prefix} are parameter-efficient methods that only require adjusting a small number of prefixes or prompt parameters, without modifying the original model's parameters. These two techniques have already been applied across various fields \cite{peng2024model,he2024prompt,fischer2024prompt,ma2024focused,wu2024apt,ma2024image,lin2023zero}. 


Inspired by these works, we propose MSuf, a novel multi-task suffix learning framework for rumor detection using time-series dual (paired) sentiments to capture dynamic sentiment information effectively. The framework contains three main modules. Given an input stream of messages, (1) an emotion LLM extracts sentiment embeddings and labels for each message-response pair, which are used to construct the time-series sequence and auxiliary task labels; (2) The Cross-Modal Fusion module aligns the temporal dual sentiment sequence with their corresponding source text to obtain aligned text-sentiment embeddings; (3) Two types of hard prompts are combined with the aligned text-sentiment stream to perform the auxiliary sentiment task and the main rumor detection task. This paper focuses on sentiment intensity (SI), which refines sentiment expression and effectively captures the differences in dual emotions as they evolve over time.

Our main contributions are:
\begin{itemize} 
\vspace*{-2mm}
\item We conduct sentiment analysis on four rumor detection datasets and design the first pipeline to mine temporally sorted dual SI. The analysis results indicate that temporal SI is strongly correlated with the veracity of rumors.
\vspace*{-2mm}
\item We propose MSuf, the first parameter-efficient multi-task suffix learning framework for rumor detection, which can also align time-series sentiments and source text. 
\vspace*{-2mm}
\item We evaluate the MSuf framework on a variety of rumor detection benchmarks. Results show that MSuf achieves significant improvements over other emotion-based methods, with the highest F1 increases of 15.3\%, 10.9\%, 23.4\%, and 15.6\% respectively. 
\end{itemize}

\section{Methodology}

\begin{figure*}[t]
  \includegraphics[width=2\columnwidth]{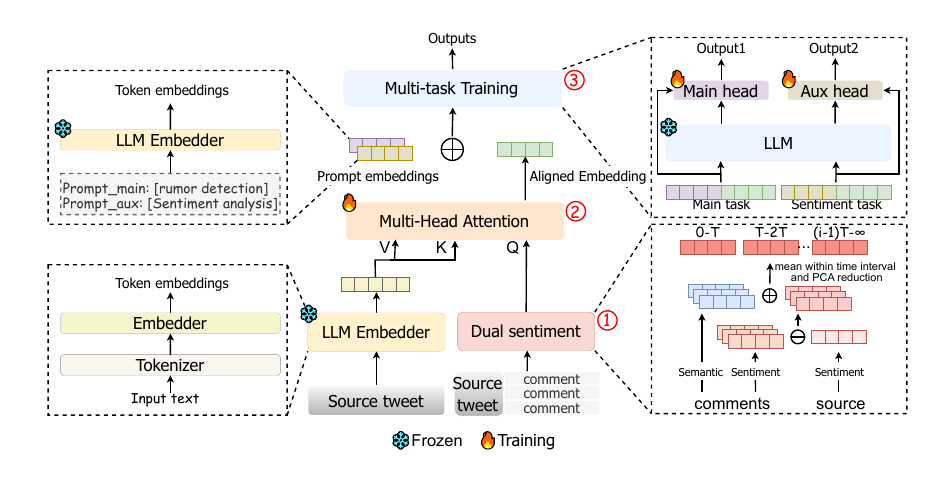}
  \caption{The architecture of MSuf. \ding{172}: Dual Sentiment Mining Module obtains sentiment intensity and semantic features, which are sorted to get the temporal dual sentiment. \ding{173}: Cross-Modal Fusion Module fuses dual sentiment and source messages to get an aligned text-sentiment embedding stream. \ding{174}: Multi-task Training Module concatenates hard prompts with the aligned embedding stream to perform multi-task suffix learning that identifies rumors. T: time interval. i: the number of intervals.}
  \label{fig:mainmethod}
\end{figure*}

This section introduces the MSuf framework for rumor detection. We first present the dataset collections and sentiment analysis in Sec \ref{sec:datasets} and Sec \ref{sec:sentimentanalysis} respectively to explore the relationship between temporal SI and rumors. Sec \ref{sec:dualemotionmining} to Sec \ref{sec:multitasktraining} introduce the three main modules of MSuf, with the overall architecture shown in Figure~\ref{fig:mainmethod}. In the \textit{dual sentiment mining} module (Sec. \ref{sec:dualemotionmining}), we apply an emotion LLM to extract SI features and sort them chronologically. The \textit{cross-modal fusion} module (Sec. \ref{sec:crossmodalfusion}) obtains the aligned embedding by fusing the dual SI stream with its source text. Finally, the \textit{multi-task training} module (Sec. \ref{sec:multitasktraining}) performs rumor detection and tracking based on one frozen LLM by multi-task suffix learning after combing two hard-prompts with the aligned embedding. 

\subsection{Datasets \label{sec:datasets}}

We collect PHEME16 \cite{zubiaga2016analysing}, PHEME18 \cite{kochkina2018all}, RumourEval19 (RE19) \cite{gorrell2019semeval}, Twitter15 \cite{ma2017detect} and Twitter16 \cite{ma2017detect} datasets. PHEME16 is a multilingual rumor dataset associated with 9 newsworthy events, containing three label values (false, true, unverified). We select the English threads in this paper. PHEME18 is an extended version of PHEME16. Each thread in PHEME18 is annotated as either rumor or non-rumor; rumors in PHEME18 are labeled as either true, false, or unverified. We combine these two-level labels by using true, false, unverified to replace the rumor label. RE19 is collected from two platforms (Twitter and Reddit), in which each news piece is labeled as true, false, or unverified. Twitter15 and Twitter16 also have four kinds of labels (true, false, unverified, non-rumor). We crawled data using conversation IDs. Due to lots of tweets being removed, we merged Twitter15 and Twitter16 and de-duplicated them into TW1516. After collecting all datasets, we remove the data items without comments. For PHEME16, PHEME18, and TW1516, we select 20\% of the data as test set and keep 20\% of the remaining data as the evaluation set. For RE19, we keep the original dataset split. Statistics of the four datasets are shown in Table \ref{tab:datasetsfour}.

\subsection{Sentiment Analysis \label{sec:sentimentanalysis}}

To determine if there is a statistical dependency between time-series dual SI signals and the actual verification judgment of the text topic/event, we create two categorical variables and conduct a chi-squared significance test. One is the misinformation label (False, True, Unverified for PHEME16 and RE19, False, True, Unverified, Non-rumor for PHEME18 and TW1516), and the other variable is temporal dual SI signals. We divide all comments into 25 intervals based on the publication time difference between the comments and source. The final interval ranges from 24*T to the end (we set T as one hour). 
We define the SI of the initial source message and the SI of its corresponding comments in each interval as $SI_{s}$ and $SI_{c}$ respectively (SI labels are all obtained by EmoLlama-chat-7b). $SI_{dual}$ = $SI_{c}$ - $SI_{s}$ measures dual sentiment differences. ``$SI_{dual>0}$: $SI_{dual}$ > 0'' means the comments are more positive, reflecting sentiment resonance and optimistic interpretations, while ``$SI_{dual<0}$: $SI_{dual}$ < 0'' means comments are more negative, reflecting sentiment amplification and negative interpretations. We count the quantities of $SI_{dual>0}$ and $SI_{dual<0}$ in each category of the rumor of 25 intervals as another variable\footnote{``$SI_{s}^{dual}$ = 0'' means complete sentiment resonance, this type is not included in the statistical process due to the small number.}. In the chi-squared statistical significance test, we start by assuming the null hypothesis, which posits that temporal dual SI signals are independent of the truthfulness of the news piece. Next, we determine whether the chi-squared statistic supports or opposes the null hypothesis. For PHEME16, the chi-squared statistic is 222.41, and the p-value is 1.32e-23, indicating that we can reject the null hypothesis. Similarly, for PHEME18, the chi-squared statistic is 3608.26, and the p-value is 0. For TW1516, the chi-squared statistic is 1064.62, and the p-value is 6.49e-139. For RE19, the chi-squared statistic is 637.74, and the p-value is 4.64e-80. In summary, the null hypothesis can be rejected across all four datasets, suggesting that temporal dual SI signals are statistically linked to the veracity of the news. 

Figure \ref{fig:temporalsentiment} in Appendix \ref{app:sichanges} presents the dual SI difference with time changes. The phenomena demonstrate that different categories of SI changes follow different trends, which can serve as important features for rumor detection. 

\subsection{Dual Sentiment Mining Module \label{sec:dualemotionmining}}

After conducting sentiment analysis and understanding the dual SI importance in rumor detection, we aim to extract the dual SI features in this module. Firstly, we apply EmoLlama-chat-7b \cite{liu2024emollms} to obtain the SI embeddings (i.e., last hidden state)\footnote{We have obtained the SI labels in the last section, which will be used in the auxiliary task.} of the source text and corresponding comments, and utilize RoBERTa to obtain the semantic embeddings (i.e., last hidden state) of message texts.

\begin{small}
    \begin{equation}
        E_s^{si} = EmoLlama (text_{source}) \in R^{1\times d^{si}}
    \end{equation}
\end{small}

\begin{small}
    \begin{equation}
        E_c^{si} = EmoLlama (text_{comments}) \in R^{num_c\times d^{si}}
    \end{equation}
\end{small}
\begin{small}
    \begin{equation}
        E_c^{sem} = RoBERTa (text_{comments}) \in R^{num_c\times d^{sem}}
    \end{equation}
\end{small}

where $E_s^{si}$ and $E_c^{si}$ denote sentiment intensity embeddings of source and comment messages respectively, and $E_c^{sem}$ denotes semantic embeddings of messages. $d^{si}$ and $d^{sem}$ are the dimension of the last hidden state of EmoLlama (4096) and RoBERTa (768). $num_c$ is the number of comments responding to the original source tweet.

\begin{small}
    \begin{equation}\label{eqn:emotioncombine}
        E_{com} = (E_c^{si} \ominus E_s^{si}) \oplus E_c^{sem} \in R^{num_c\times (d^{si} + d^{sem})}
    \end{equation}
\end{small}

\begin{small}
    \begin{center}
    \begin{equation}\label{eqn:combinesort}
    \begin{aligned}
        E_{sort}  =  & sort (E_{com}) \\
        =& \{ [e_{0}^{0-T},e_{1}^{0-T},...],[e_{0}^{T-2T},e_{1}^{T-2T},...],... \\
        &[e_{0}^{(i-1)T-\infty},e_{1}^{(i-1)T-\infty},...] \}
    \end{aligned}
    \end{equation}
    \end{center}
\end{small}

We obtain the fusion embeddings $E_{com}$ as shown in Equation \ref{eqn:emotioncombine}. Subtraction between source SI $E_s^{si}$ and subsequent comment SI $E_c^{si}$ produces the SI fusion vector, which will be concatenated with semantic embeddings $E_c^{sem}$. We sort the embeddings $E_{com}$ by time interval to construct the final fusion embeddings $E_{sort}$\footnote{The posting time of each source is denoted as 0. The time of comments discussed in this paper occurs after the source message publication.}. Here, $i$ represents the number of time intervals, and $T$ represents the size of each time interval. The final interval consists of all comments posted after $(i-1)T$ time points. 

\begin{small}
    \begin{equation}
        \begin{aligned}
        E_{ave} &= ave (E_{sort}) \\
        &= \{e_{t_0}^{ave},e_{t_1}^{ave},...e_{t_{i-1}}^{ave} \} \in R^{i\times (d^{si} + d^{sem})}
        \end{aligned}
    \end{equation}
\end{small}

\begin{small}
    \begin{equation}
        E_{dual} = PCA (E_{ave}) \in R^{i\times d^{dual}}
    \end{equation}
\end{small}

We calculate the average of embeddings within each time interval to obtain $E_{ave}$. Finally, we adopt the PCA technique to reduce the original dimensions ($d^{si} + d^{sem}$) to $d^{dual}$ dimensions.

\subsection{Cross-Modal Fusion Module \label{sec:crossmodalfusion}}

In this module, we attempt to align the time series dual sentiment $E_{dual}$ obtained in Sec \ref{sec:dualemotionmining} and the LLM word embedding of the source message. We first obtain the word embedding representation of source message through LLM, followed by Multi-head Self Attention (MHSA) to get the projected dual sentiment tokens $E_{time}$.

\begin{small}
    \begin{equation}
        E_s^{word} = LLMEmbedder (text_{source}) \in R^{num_{w}\times d^{llm}}
    \end{equation}
\end{small}

\begin{small}
    \begin{equation}
        E_{time} = MHSA (MAP((E_{dual})^{Trans})) \in R^{d^{dual}\times d^{llm}}
    \end{equation}
\end{small}

Where $E_s^{word}$ is the word embedding of source text. $LLMEmbedder$ denotes the LLM embedder. $num_{w}$ is the number of tokens, $d^{llm}$ denotes the representation dimension of the LLM. The MAP layer MAP(·) performs a channel-wise dimensional mapping from $i$ to $d^{llm}$ \cite{Liu2024CALFAL}.

After this, we apply Multi-head Cross-Attention with $E_s^{word}$ as key (K) and value (V), and $E_{time}$ as query (Q) to align the source word embeddings and temporal dual SI tokens to obtain the temporally sequenced aligned text-sentiment embedding stream $E_{align}$,

\begin{small}
    \begin{equation}
        \begin{array}{c}
        E_{align} =\operatorname{Softmax}\left(\frac{Q K^{T}}{\sqrt{num_{w}}}\right) V \in R^{d^{dual}\times d^{llm}}, \\ \\
        Q=E_{time} W_{q}, K=E_s^{word} W_{k}, V=E_s^{word} W_{v},
        \end{array}
    \end{equation}
\end{small}

where $W_{q}$, $W_{k}$ and $W_{v}$ $\in$ $R^{d^{llm}\times d^{llm}}$ are the projection matrices for Q, K, and V, respectively.

\subsection{Multi-task Training Module \label{sec:multitasktraining}}


While prompting effectively activates and enhances the performance of LLMs on specific tasks \cite{chang2024efficient}, transforming the fused dual SI features into natural language and integrating them with prompts remains a significant challenge. Recent studies have demonstrated that integrating diverse data modalities (e.g., images, time series) with prompts can substantially improve the adaptability of LLMs to downstream tasks \cite{tsimpoukelli2021multimodal, jin2023time}. Inspired by these findings, we propose a multi-task suffix learning structure to facilitate rumor detection based on temporal dual SI information. The learned suffixes provide the ability to identify and track rumor messages in the message stream. 

We design prompts for both the main task (rumor detection) and the auxiliary task (SI analysis, we adopt the SI labels from Emollama as gold labels). The prompt templates are shown in Appendix \ref{app:promptexamplestasks}. We apply LLMEmbedder to encode the two prompts to obtain the prompt embeddings, which are concatenated with the aligned text-sentiment embeddings described earlier.

\begin{small}
    \begin{equation}
        E_{main} = LLMEmbedder(Prompt_{main}) \oplus E_{align}
    \end{equation}
\end{small}

\begin{small}
    \begin{equation}
        E_{aux} = LLMEmbedder(Prompt_{aux}) \oplus E_{align}
    \end{equation}
\end{small}

where $E_{main}$ and $E_{aux}$ are the combined embeddings of the main task prompt and auxiliary prompt with aligned text-sentiment embeddings. After that, the fused embeddings serve as the input to the frozen LLM to produce outputs, where a residual structure is used to add the original input. Subsequently, the outputs of the main task $output_{main}$ and the auxiliary task $output_{aux}$ can be obtained through two linear layers respectively.

\begin{small}
    \begin{equation}
        output_{main} = liner_{main}({LLM(E_{main}) + E_{main}})
    \end{equation}
\end{small}

\begin{small}
    \begin{equation}
        output_{aux} = liner_{aux}({LLM(E_{aux}) + E_{aux}}) 
    \end{equation}
\end{small}

For the main task, we apply cross-entropy loss ($loss_{main}$) as the objective function. For the auxiliary task, we adopt mean-squared-error loss ($loss_{aux}$) as the objective function. The total loss can be calculated as follows:

\begin{small}
    \begin{equation}
        loss_{total} = loss_{main} + \lambda \cdot loss_{aux}
    \end{equation}
\end{small}

where $\lambda$ is a hyperparameter that controls the magnitude of the auxiliary loss.

\section{Experiments}

\subsection{Experimental settings}

Base model: GPT-2 \cite{Radford2019LanguageMA}, Llama3.2-1b\footnote{https://www.llama.com/} (Llama).
We use one Nvidia V100 GPU (16G) to perform experiments. All models are trained for ten epochs based on AdamW optimizer \cite{loshchilov2017decoupled4}, utilizing early stopping techniques \cite{dodge2020fine1} to prevent overfitting (tolerance is set to 3). For learning rate (lr) and batch size (bs), we perform a grid search strategy \cite{liashchynskyi2019grid} with lr: [1e-2, 1e-3, 1e-4], bs: [4, 8, 16, 32]. The final results are the average of three different seed runs. 

In the following experiments, we set the time interval ($T$) to 1 hour, the number of intervals ($i$) to 25, dual dimensions ($d^{dual}$) to 256, and $\lambda$ to 0.5.  The selection process of hyperparameters can be found in Appendix \ref{selectionofhyperparas}.

\subsection{Baseline models}

We make comparisons with a series of baseline methods as follows:

\begin{itemize} 
\small

\item LLMs\footnote{We also evaluate GPT-2 and Llama3.1-1b-instruct using the zero-shot prompt method, but they cannot give related answers and get low scores.}: We apply GPT-3.5-turbo-0125 (GPT-3.5), GPT-4o-mini, GPT-4o \cite{achiam2023gpt}.

\item RAEmoLLM \cite{liu2024raemollm} is a framework for misinformation detection by retrieval augmenting demonstrations through affective information. We choose the best-performing LLM (GPT-3.5) from the above as the base model. For comparison, we apply SI for retrieval.

\item RoBERTa \cite{liu2019roberta7}: One pre-trained language model, which is commonly used for classification tasks.

\item MDE \cite{zhang2021mining} applies dual emotional features to represent the relationship between the emotions of the source and the response message, and combines them for rumor detection. The base model is BiGRU.

\item BiGCN \cite{bian2020rumor} adopt bi-directional graph model to learn the patterns of rumor propagation. 

\item BiGRU \cite{zhang2021mining}: One text-based model. We use GloVe \cite{pennington2014glove} for word embeddings. The max sequence length of BiGRU is 100, and the dimensionality of the hidden state of BiGRU is 32.



\item We also compare our method with BERT, NileTMRG, Emoratio, EmoCred, AGAT, Double-Channel, S2MTL-FND in RE19. The results refer to \cite{zhang2021mining,jiang2024makes}. The specific descriptions can be found in Appendix \ref{app:baselines}. 

\end{itemize} 

For the baselines, we conduct two kinds of experiments: one based only on the source content (marked as sou-only) and the other on a simple concatenation of the source content and the corresponding comments (without special marks). The template for simple concatenation: \textit{Source text: [text], Corresponding Comments: [comments list]}.

\subsection{Evaluation Metric}

Rumor detection is typically regarded as a classification task. We employ metrics Accuracy, Precision, Recall, and macro-F1 for evaluation \cite{su2020motivations3}.

\subsection{Results}

\begin{table*}[!t]
\footnotesize
\resizebox{\textwidth}{!}{
\begin{tabular}{lcccccccccccc}
\hline
                         & \multicolumn{4}{c}{PHEME16}                                       & \multicolumn{4}{c}{PHEME18}                                                   & \multicolumn{4}{c}{TW1516}                                        \\
\multirow{-2}{*}{Method} & ACC            & PRE            & REC            & F1             & ACC                  & PRE                  & REC            & F1             & ACC            & PRE            & REC            & F1             \\ \hline
RoBERTa-sou-only                                   & 0.567          & 0.387          & 0.428          & 0.364          & 0.779                & 0.690                & 0.642          & 0.657          & 0.732          & 0.669          & 0.662          & 0.634          \\
RoBERTa                                         & 0.528          & 0.249          & 0.390          & 0.296          & 0.770                & 0.714                & 0.610          & 0.635          & 0.758          & 0.708          & 0.700          & 0.688          \\
BiGRU                                           & 0.617          & 0.619          & 0.555          & 0.562          & 0.775                & 0.679                & 0.661          & 0.665          & 0.570          & 0.493          & 0.518          & 0.499          \\
BiGCN                                           & \underline{0.672}    & 0.649          & \underline{0.637}    & \underline{0.639}    & \underline{\textbf{0.808}} & \underline{\textbf{0.763}} & 0.672          & \underline{0.707}    & \underline{0.775}    & \underline{0.761}    & \underline{0.749}    & \underline{0.754}    \\
MDE                                             & 0.572          & 0.591          & 0.524          & 0.537          & 0.759                & 0.650                & 0.599          & 0.611          & 0.646          & 0.578          & 0.584          & 0.566          \\ \hline
GPT-4o-mini                                     & 0.350          & 0.278          & 0.368          & 0.224          & 0.475                & 0.408                & 0.381          & 0.284          & 0.403          & 0.536          & 0.388          & 0.322          \\
GPT-4o                                          & 0.417          & 0.456          & 0.407          & 0.346          & 0.608                & 0.232                & 0.241          & 0.229          & 0.451          & 0.279          & 0.272          & 0.266          \\
GPT-3.5-sou-only                                   & 0.550          & 0.360          & 0.464          & 0.404          & 0.576                & 0.417                & 0.449          & 0.422          & 0.457          & 0.417          & 0.388          & 0.388          \\
GPT-3.5                                         & 0.483          & 0.358          & 0.446          & 0.361          & 0.216                & 0.341                & 0.322          & 0.206          & 0.273          & 0.391          & 0.305          & 0.285          \\
RAEmoLLM                                        & 0.567          & \underline{0.702}    & 0.532          & 0.522          & 0.726                & 0.681                & 0.622          & 0.601          & 0.672          & 0.677          & 0.658          & 0.647          \\ \hline
MSuf                                            & \textbf{0.700} & \textbf{0.720} & \textbf{0.666} & \textbf{0.675} & 0.797                & 0.737                & \textbf{0.695} & \textbf{0.710} & \textbf{0.827} & \textbf{0.808} & \textbf{0.800} & \textbf{0.800} \\
Msuf-Llama                                      & 0.678          & 0.675          & 0.674          & 0.667          & 0.785 & 0.695 & 0.689 & 0.691 & 0.795 & 0.768 & 0.769 & 0.767        \\ \hline
\end{tabular}
}
\caption{\label{tab:results}
Results on PHEME16, PHEME18, and TW1516. Those without special markers indicate using both source and comments and ``sou-only'' means using only source messages. Bold indicates best performance. Underlining denotes the best performance among the baselines excluding MSuf and MSuf-Llama. For GPT-4 series and RAEmoLLM, we only present the best version from ``sou-only'' and using both source and comments.}
\end{table*}

\begin{table}[htb]
\footnotesize
\centering
\begin{tabular}{p{5cm}cc}
\hline
{\color[HTML]{212121} Method} & \multicolumn{2}{c}{RumourEval19} \\
                              & ACC             & F1             \\ \hline
BiGRU                         & -               & 0.269          \\
MDE                           & -               & 0.346          \\
NileTMRG                      & -               & 0.309          \\
Emoratio \cite{ajao2019sentiment}                     & -               & 0.271          \\
EmoCred \cite{giachanou2019leveraging}                      & -               & 0.308          \\
AGAT \cite{li2021joint}                          & -               & 0.370          \\
BERT                          & -               & 0.272          \\
RoBERTa                       & 0.300          & 0.193         \\ 
Double-Channel \cite{kim2023detecting}               & 0.494           & 0.403          \\
S2MTL-FND \cite{jiang2024makes}                    & \textbf{0.519}  & \underline{0.414}    \\ \hline
GPT-4o-mini                   & 0.247           & 0.214          \\
GPT-4o                        & 0.309           & 0.262          \\
GPT-3.5-sou-only                 & 0.358           & 0.354          \\
GPT-3.5                       & 0.210           & 0.186          \\
RAEmoLLM                      & 0.296           & 0.295          \\
MSuf                          & \underline{0.502}     & \textbf{0.427} \\
Msuf-Llama                    & 0.469           & 0.392          \\ \hline
\end{tabular}
\caption{\label{tab:results2}
Results on RE19 dataset. Bold indicates best performance. Underlining denotes the second-best performance. In the results of the first column, all except RoBERTa are referenced from \cite{zhang2021mining,JIANG2024112395}.}
\end{table}

Table \ref{tab:results} and \ref{tab:results2} present the results on four datasets of different baseline methods and MSuf. MSuf applies GPT-2 as the base model. It achieves SOTA F1 performance across four datasets. Next, we analyze the experimental results in detail (discussing results exclusively based on F1 score).

\textbf{Comparsion with traditional models:} In the datasets with smaller amounts of data (PHEME16 and RE19), we can see that RoBERTa and Bi-GRU perform largely worse compared to the MSuf, but show significant improvement in PHEME18 and TW1516. This indicates that series models like RoBERTa and Bi-GRU require a certain amount of data to train effectively, which also illustrates that our proposed MSuf method can effectively learn the necessary features on small datasets.

\textbf{Comparsion with LLMs:} MSuf method performs better than LLMs among all datasets. The LLMs with zero-shot prompts are worse than the fine-tuned models in most cases except for RE19. RE19 is collected from different platforms. GPT-3.5 performs well (GPT-4o series still performs worse\footnote{One possible reason is that the GPT-4o series is more advanced than GPT-3.5, with stronger generation capabilities, which may result in a decrease in its classification ability.}). We can also observe that, after adding comment messages, the performance of most LLMs decreases, which indicates that simply concatenating source text with its comments is not an effective combination method. 
Overall, the MSuf method outperforms LLMs in terms of overall performance and can effectively utilize both source and comment information.

\textbf{Comparison with emotion/sentiment-based methods:} MDE is one typical method that fuses dual emotion and RAEmoLLM is a framework to improve LLMs through retrieval using SI embeddings. In this paper, we apply GPT-3.5 as the base model for RAEmoLLM since it has the best performance among LLMs. RAEmoLLM improves the performance of GPT-3.5 largely in PHEME16, PHEME18 and TW1516 datasets. However, these emotion/sentiment-based methods, including the NileTMRG, Emoratio, EmoCred, S2MTL-FND, all perform worse than the MSuf method. This indicates that MSuf can effectively and fully utilize SI features.

\textbf{Comparison between comment-based methods:} By integrating features of comments (e.g. NileTMRG, Double-Channel), or constructing a graph structure to simulate rumor propagation (e.g. AGAT, BiGCN), can achieve better results. Double-Channel and BiGCN perform similarly to MSuf, but MSuf still maintains an advantage, further highlighting the effectiveness of incorporating SI features.

\textbf{Comparison between different base models for MSuf:} From Table \ref{tab:results}, we can see MSuf based on GPT-2 is better than MSuf-Llama in all datasets. One possible reason is that the latest LLMs are more focused on generation, so using an external fully connected layer for classification may affect their effectiveness. But our MSuf framework can still largely enhance the performance of original LLMs.

Overall, MSuf is an effective framework for enhancing model performance, capable of fully utilizing dual SI information.

\subsection{Ablation study}

\begin{table}[!t]
\centering
\footnotesize
\resizebox{0.48\textwidth}{!}{
\begin{tabular}{lcccccc}
\hline
                                                & \multicolumn{2}{c}{PHEME16}     & \multicolumn{2}{c}{TW1516}      & \multicolumn{2}{c}{RE19} \\
\multirow{-2}{*}{Method} & ACC            & F1             & ACC            & F1             & ACC             & F1             \\  \hline
Msuf                                            & \textbf{0.700} & \textbf{0.675} & \textbf{0.827} & \underline{0.800}    & \textbf{0.502}  & \textbf{0.427} \\
Full-para                                       & 0.650          & 0.615          & 0.811          & 0.780          &     0.453            &       0.376         \\
Msuf-Llama                                      & 0.678          & 0.667          & 0.795         & 0.767         & 0.469           & 0.392          \\ \hline
                                                & \multicolumn{6}{c}{Dual Emotion Mining Module}                                                       \\
w/o semantic                                    & 0.678          & \underline{0.657}    & 0.807          & 0.776          & 0.490           & 0.389          \\
w/o sentiment                                   & 0.667          & 0.631          & 0.816          & 0.779          & \underline{0.494}     & \underline{0.400}    \\
w/o sou sentiment                               & 0.672          & 0.633          & 0.818          & 0.790          & 0.477           & 0.374          \\
com sentiment-only                                & 0.672          & 0.632          & \underline{0.826}    & \textbf{0.801} & 0.481           & 0.390          \\
w/o time                                        & 0.672          & 0.644          & 0.815          & 0.781          & 0.465           & 0.389          \\ \hline
                                                & \multicolumn{6}{c}{Cross-Modal fusion Module}                                                        \\
w/o align                                       & 0.600          & 0.515          & 0.623          & 0.494          & 0.412           & 0.340          \\ \hline
                                                & \multicolumn{6}{c}{Multi-task Training Module}                                                       \\
align first                                     & 0.611          & 0.511          & 0.820          & 0.787          & 0.358           & 0.321          \\
align-only                                        & \underline{0.683}    & 0.649          & 0.823          & 0.790          & 0.444           & 0.332          \\
w/o aux                                         & 0.650          & 0.611          & 0.815          & 0.788          & 0.465           & 0.370          \\
w/o residual                                    & 0.611          & 0.566          & 0.817          & 0.787          & 0.481           & 0.337          \\ \hline
\end{tabular}
}
\caption{\label{tab:ablationanalysis}
Ablation Analysis. ``w/o'' denotes without. ``sou'' denotes source. ``com'' denotes comments. Bold indicates best performance. Underlining denotes the second-best performance.}
\end{table}

To verify the validity of each module of the MSuf framework, we design multiple different versions by removing different factors for ablation experiments as follows:

{\small (1) For the dual sentiment mining module.}
\begin{itemize}
\small

    \item w/o semantic: Do not use semantic embeddings $E_c^{sem}$.
    \item w/o sentiment: Do not use SI embeddings $E_c^{si}$ and $E_s^{si}$.
    \item w/o sou sentiment: Do not use source SI embeddings $E_s^{si}$.
    \item com sentiment-only: Only use comment SI embeddings $E_c^{si}$
    \item w/o time: Do not use temporal sequence, just randomly split the comments into 25 intervals.
\end{itemize}

{\small (2) For Cross-Modal Fusion Module.}
\begin{itemize}
    \small
    \item w/o align: Do not use aligned text embedding. Just concatenate the prompt and the source text.
\end{itemize}

{\small (3) For the Multi-task Training Module.}
\begin{itemize}
    \small
    \item align first: Concatenate aligned embedding and prompt in [align, prompt] order. The prompt is adjusted accordingly.
    \item align-only: Only use aligned embedding without prompts.
    \item w/o aux: Do not use the sentiment auxiliary task.
    \item w/o residual: Do not use the residual structure.
\end{itemize}

Table \ref{tab:ablationanalysis} shows the results of the ablation experiment. Different model variants all exhibit varying degrees of performance decrease except \textit{com sentiment-only} variant in TW1516\footnote{This may be related to the TW516 dataset having rich comments (an average of about 21 comments per source tweet). In contrast, the average number of comments in other datasets is as follows: PHEME16: 16, PHEME18: 17, RE19: 18.}. The results of the \textit{without-semantic} variant indicate that semantic information can enhance the robustness of MSuf, especially for the multi-source RE19 dataset. Over the three datasets, the \textit{without sentiment} variant decreased by 4.4\%, 2.1\%, and 2.7\%, respectively. This highlights the importance of SI information in dual sentiment mining. The performance reduction in the \textit{without-time} variant confirms the effectiveness of temporal features. The \textit{without-align} variant results illustrate using only the source content is not ideal and the cross-modal fusion module can produce meaningful combination information of temporal dual SI features and source content. The decreased results of the \textit{align-first} variant and \textit{align-only} variant demonstrate our proposed suffix structure (i.e., hard prompt + aligned text embedding) is the most effective, especially in the RE19 dataset. The results from \textit{without-aux} variant indicate that the sentiment auxiliary task helps the main task. Full-para is a variant with LLM parameters unfrozen, which is not better than MSuf, especially at small scales, indicating that MSuf can effectively avoid overfitting or dropping in local optima.

\subsection{Early Detection Analysis}

\begin{figure}[htb]
  \includegraphics[width=\columnwidth]{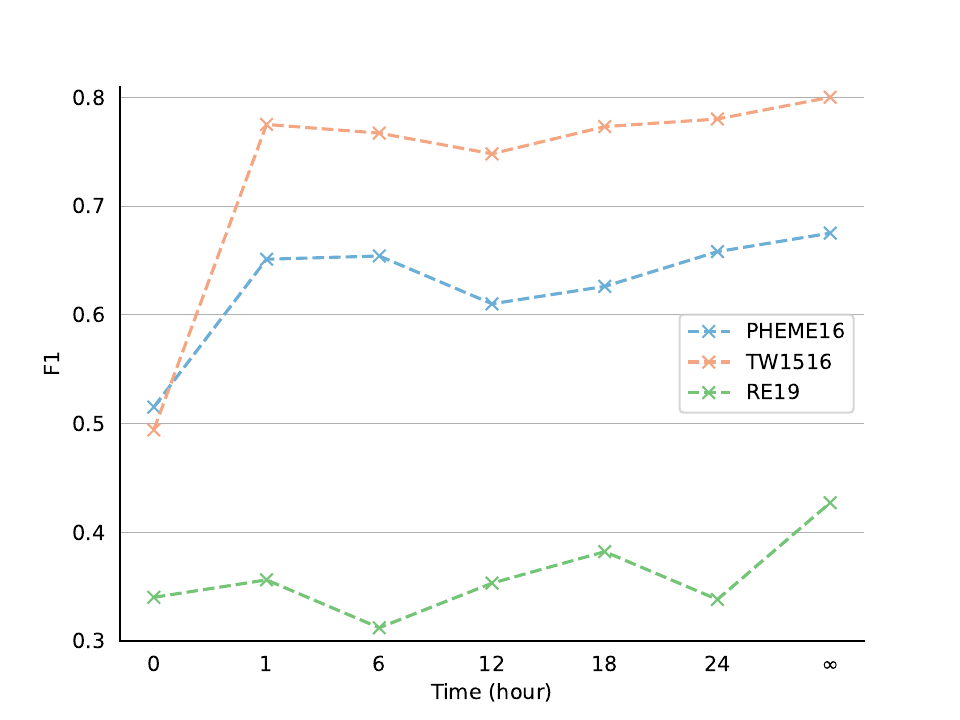}
  \caption{Early detection results on three datasets}
  \label{fig:earlyrumordetection}
\end{figure}

To assess the real-time performance and effectiveness of MSuf in rumor detection, we conduct experiments based on the different number of intervals. We chose seven detection time points (0 (i.e., without-align), 1h, 6h,  12h, 18h, 24h, and $\infty$ (i.e., Full MSuf)). Figure \ref{fig:heatmap_fourdatasets} shows that most source messages receive many replies in the first 1 hour after posting. This can explain why the performance using the comments in the first 1 hour is close to the full MSuf in the PHEME16 and TW1516 datasets in figure \ref{fig:earlyrumordetection}. However, for the complex RE19 data, using fewer comments seems not ideal, it is better to collect as many as possible corresponding comments. In conclusion, for datasets from a single data source (e.g., PHEME16 and TW1516), the MSuf method can achieve good rumor detection performance within the first 1 hour. However, for datasets collected from multiple sources (e.g., RE19), MSuf needs more comment data to confirm rumors.

\subsection{Generalization Evaluation}

\begin{figure}[htb]
  \includegraphics[width=\columnwidth]{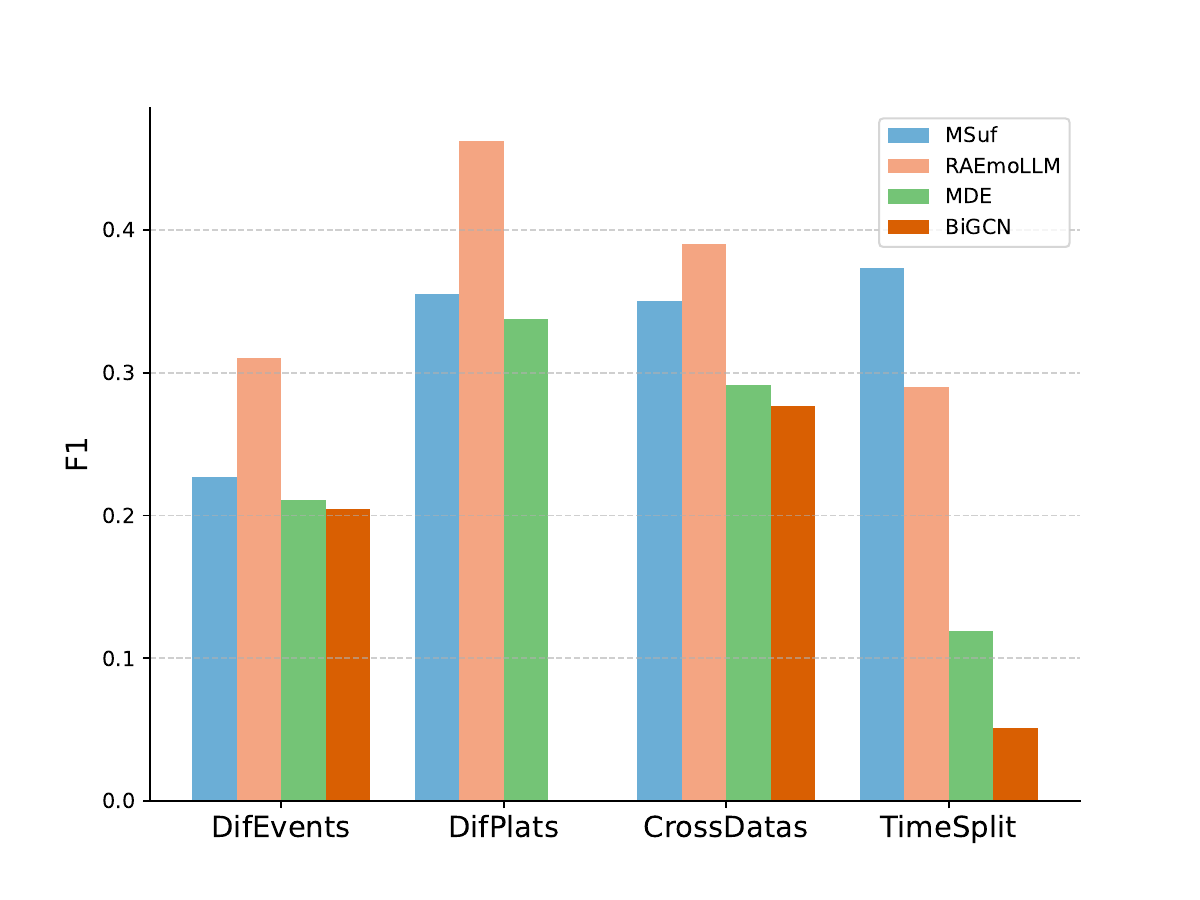}
  \caption{Generalization evaluation on different settings. ``DifEvents'' denotes testing on data from different events. ``DifPlats'' denotes testing on data from different platforms. ``CrossDatas'' denotes testing across datasets. ``TimeSplit'' denotes testing on data occurring late.}
  \label{fig:generalizationtests}
\end{figure}

To evaluate the generalization ability of MSuf, we set four experiments. 

(1) Evaluation on data from different events (DifEvents). We use PHEME18 and set Ferguson, Prince, and Ebola events as the test set, Ottawashooting, Putinmissing, and Gurlitt events as the validation set, and the remaining three events as the training set.

(2) Evaluation on data from different platforms (DifPlats). We divide RE19 into Twitter and Reddit parts. The Reddit part is used as the test set, while the Twitter part is used for the training and validation sets. 

(3) Evaluation across datasets (CrossDatas). We apply PHEME18 as training and validation sets, while use TW1516 as test set.

(4) Evaluation on data occurring late (TimeSplit). We sort the TW1516 data by the publication time of the source tweet, selecting the last 300 items as the test set, the last 300 items from the remaining data as the validation set, and the rest of the data as the training set.

Figure \ref{fig:generalizationtests} presents the performance on different settings. Due to not requiring fine-tuning, RAEmoLLM is not significantly affected by these settings. Among the remaining methods, MSuf performs the best, followed by MDE, and BiGCN performs the worst. This confirms the importance of utilizing SI information.

\section{Conclusion}

In this paper, we propose the MSuf framework, the first multi-task suffix learning framework for rumor detection using temporally sorted dual (source-comments coupled) sentiment intensity. We introduce three modules of MSuf. We also conduct a comprehensive sentiment analysis for four public rumor datasets to explain the relationship between temporal dual SI and rumors. We evaluate the performance of MSuf on the four rumor detection benchmarks. The results show that MSuf can significantly improve LLMs compared to other emotion-based methods, comment-based methods and zero-shot LLMs. We also conduct an ablation analysis of each module, early detection analysis and four generalization evaluation experiments, which provide a foundation for further improvements.

In the future, we will explore the application of the MSuf framework for multi-lingual and multi-modal misinformation tasks. We will also evaluate the application of the MSuf framework in other fields. There are many other factors influencing misinformation, such as stance and topic. We will combine sentiments with other features to construct a more robust MSuf framework.

\section{Limitations}

Due to restricted computational resources, we only applied GPT-2, Llama3.2-1b open-sourced LLMs as the base model to evaluate MSuf framework. As such, we have not considered how the use of larger or different model architectures may potentially impact upon performance in rumor detection tasks.

The results show that the MSuf framework based on LLama does not surpass the performance of MSuf based on GPT-2. A possible reason is that current popular LLMs are more focused on generative capabilities, and adding an external classification layer may impact their performance. Therefore, optimizing the framework to better align with recent LLMs is worth further exploration.

We found that many data points were no longer available when collecting the Twitter15 and Twitter16 datasets due to the long time passed, potentially compromising conversation completeness.

Due to dataset limitations, the debunking time of rumors and whether users obtain information from external platforms remain uncertain. This may affect the accuracy of sentiment analysis to some extent.


\section{Ethics Statement}

The datasets we use in this paper are sourced from public social media platforms and websites. We strictly adhere to privacy agreements and ethical principles to protect user privacy and to ensure the proper application of anonymity in all texts.



\bibliography{custom}


\appendix

\section{Related Work}

\subsection{Temporal Misinformation Detection}

Posts on social media often come with timestamps. Therefore, many studies use time information to assist in detecting false information. \citet{nie2021early} propose a temporal Bi-GCN model to learn representations for rumor propagation by encoding the temporal information for graph structures. \citet{qu2024temporal} design multimodal graph neural networks that can form a temporal news cluster and learn temporal features for fake news detection. \citet{guo2022temporal} suggest a block-based representation and fusion mechanism that can utilize the information from both spatial and temporal perspectives. \citet{wakamiya2020fake} adopt an attention-based method to combine linguistic and user features alongside temporal features. \citet{cavalcante2024early} propose the time-aware crowd signals method to explore the temporal nature of news propagation, utilizing users’ reputations obtained from their public behavior when spreading news in the past.



\subsection{Emotion-based Misinformation Detection}

Emotion and sentiment are important features in detecting misinformation \cite{liu2024emotion}. 
\citet{choudhry2022emotion,chakraborty2023emotion,jiang2024makes,kumari2024emotion} introduce multitask frameworks for incorporating sentiment/emotion as auxiliary tasks to enhance the main misinformation detection task. \citet{ghanem2021fakeflow} propose the FakeFlow for fake news detection that can capture the flow of affective information by combining topic and sentiments. \citet{Liu2024ConspEmoLLMCT,liu2024raemollm} adapt LLMs for misinformation detection by levering affective information based on fine-tuning and RAG techniques respectively. However, these models are only applicable to fake source text and do not consider social content. \citet{luvembe2023dual} develop an attention-based mechanism by combining CNN and Bi-GRU for enriched extraction of dual emotion features. \citet{zhang2021mining} analyze the relationship between dual emotions and propose dual emotion features, which can be plugged into existing fake news detectors. \citet{wang2024multimodal} further fuses the author's visual emotion for multi-modal rumor detection. These works indicate the importance of dual emotion features. 

\subsection{Comment-based Rumor Detection}

Social media posts often come with rich comments. Combining information from the source and tweets can help improve rumor detection models. Some studies focus on the propagation pattern in rumors. \citet{ma-etal-2018-rumor} apply two recursive neural models based on tree-structured neural networks to learn the propagation layout of tweets. \citet{bian2020rumor} adopt a bi-directional graph model to learn the patterns of rumor propagation. \citet{khoo2020interpretable} propose a post-level attention model, which incorporates tree structure information into the transformer network.  \cite{yang-etal-2023-rumor} design a Crowd Intelligence and ChatGPT-Assisted Network, which combines crowd intelligence-based feature learning, ChatGPT-enhanced knowledge mining, and an entity-aware heterogeneous graph. Also, there are many studies using the features of comments. Like \citet{zhang2021mining} apply the emotion and sentiment features and \citet{enayet2017niletmrg,li2021joint} leverage stance features for misinformation detection.



\subsection{Prompt-tuning and Prefix-tuning}

Prompt-tuning \cite{lester2021power} and prefix-tuning \cite{li2021prefix} typically involve fine-tuning a small number of parameters to adapt to specific tasks, rather than modifying the entire LLM. \citet{peng2024model} develop a soft prompt-based learning architecture based on a clinical LLM for clinical concept and relation extraction. \citet{fischer2024prompt} provide a prompt-able architecture for medical image segmentation, which is frozen post pre-training but remains flexible with class-specific learnable prompt tokens. \citet{ma2024image} propose a super-resolution training method based on prefix and prompt fine-tuning, in which only a few prefix and prompt parameters are added to the self-attention module. \citet{wu2024apt} design an adaptive prefix-tuning technique involving the training of prefix parameters on adapting tasks, followed by fine-tuning on downstream tasks. \citet{ma2024focused} propose focused prefix tuning to enable the control to focus on the desired attribute for controllable text generation. \citet{lin2023zero} propose a zero-shot framework for rumor detection based on prompt learning. They first represent Rumors as different propagation threads, then a hierarchical prompt encoding mechanism is designed to learn context representations.

\begin{figure}[htb]
\centering
  \includegraphics[width=0.9\columnwidth]{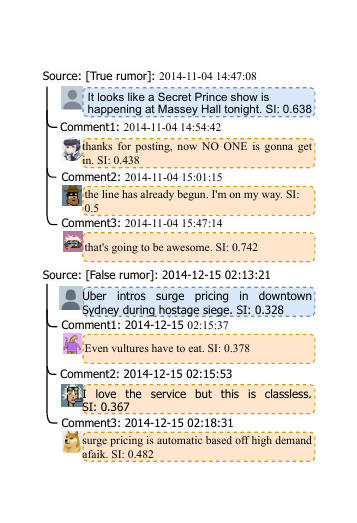}
  \caption{Rumor examples.}
  \label{fig:rumorsamples}
\end{figure}

\section{Rumor samples (Figure \ref{fig:rumorsamples})}

\section{Data Statistic (Table \ref{tab:datasetsfour})}

\begin{table*}[htb]
\centering
\small
\begin{tabular}{cccccccccc}
\hline
{\color[HTML]{212121} }                        &                            & \multicolumn{2}{c}{PHEME16} & \multicolumn{2}{c}{PHEME18} & \multicolumn{2}{c}{RumourEval19} & \multicolumn{2}{c}{TW1516} \\
\multirow{-2}{*}{{\color[HTML]{212121} Split}} & \multirow{-2}{*}{Veracity} & source      & comments      & source      & comments      & source         & comments        & source      & comments     \\ \hline
                                               & FALSE                      & 40          & 577           & 326         & 3854          & 79             & 1135            & 216         & 2749         \\
                                               & TRUE                       & 85          & 1438          & 605         & 9092          & 143            & 1905            & 224         & 2202         \\
                                               & Unverified                 & 64          & 1112          & 396         & 6944          & 104            & 1838            & 149         & 2518         \\
                                               & Non-Rumor                  & -           & -             & 2351        & 43440         & -              & -               & 345         & 11974        \\
\multirow{-5}{*}{Training}                     & Total                      & 189         & 3127          & 3678        & 63330         & 326            & 4878            & 934         & 19443        \\ \hline
                                               & FALSE                      & 9           & 76            & 72          & 833           & 19             & 824             & 51          & 622          \\
                                               & TRUE                       & 23          & 249           & 169         & 2072          & 10             & 404             & 55          & 523          \\
                                               & Unverified                 & 16          & 341           & 94          & 1534          & 9              & 212             & 36          & 645          \\
                                               & Non-Rumor                  & -           & -             & 585         & 10930         & -              & -               & 92          & 3111         \\
\multirow{-5}{*}{Validating}                   & Total                      & 48          & 666           & 920         & 15369         & 38             & 1440            & 234         & 4901         \\ \hline
                                               & FALSE                      & 13          & 156           & 107         & 1224          & 40             & 689             & 66          & 744          \\
                                               & TRUE                       & 29          & 366           & 198         & 3044          & 31             & 805             & 73          & 602          \\
                                               & Unverified                 & 18          & 481           & 127         & 2126          & 10             & 181             & 43          & 600          \\
                                               & Non-Rumor                  & -           & -             & 718         & 13836         & -              & -               & 111         & 3728         \\
\multirow{-5}{*}{Testing}                      & Total                      & 60          & 1003          & 1150        & 20230         & 81             & 1675            & 293         & 5674         \\ \hline
                                               & FALSE                      & 62          & 809           & 505         & 5911          & 138            & 2648            & 333         & 4115         \\
                                               & TRUE                       & 137         & 2053          & 972         & 14208         & 184            & 3114            & 352         & 3327         \\
                                               & Unverified                 & 98          & 1934          & 617         & 10604         & 123            & 2231            & 228         & 3763         \\
                                               & Non-Rumor                  & -           & -             & 3654        & 68206         & -              & -               & 548         & 18813        \\
\multirow{-5}{*}{Total}                        & Total                      & 297         & 4796          & 5748        & 98929         & 445            & 7993            & 1461        & 30018        \\ \hline
\end{tabular}
\caption{\label{tab:datasetsfour}
Statistic of datasets.}
\end{table*}

\section{The selection process of hyperparameters \label{selectionofhyperparas}}

\begin{figure*}[htbp]
    \centering
    \subfloat[PHEME18]{\includegraphics[width=1\textwidth] {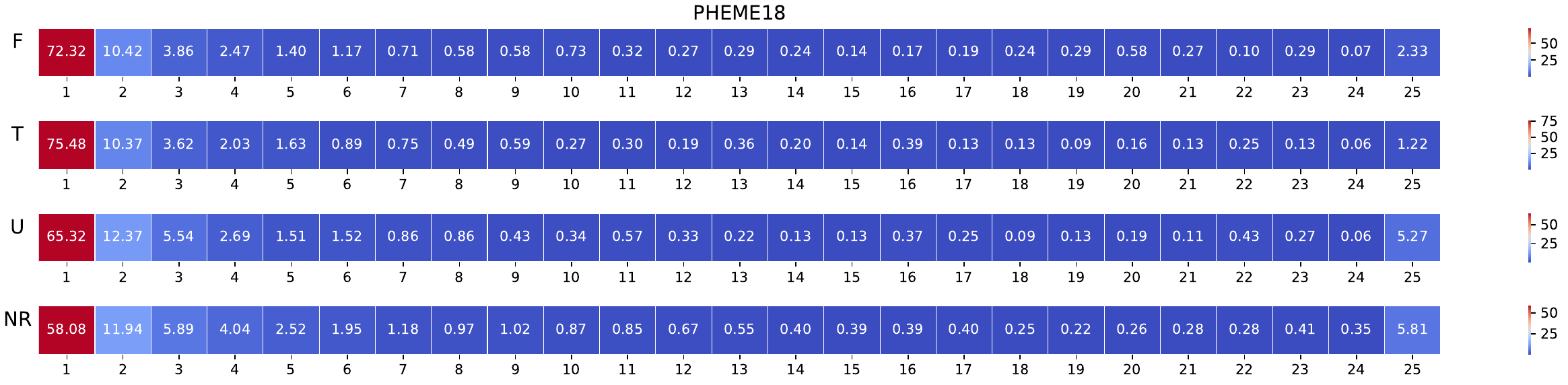}} \\
    \subfloat[TW1516]{\includegraphics[width=1\textwidth]{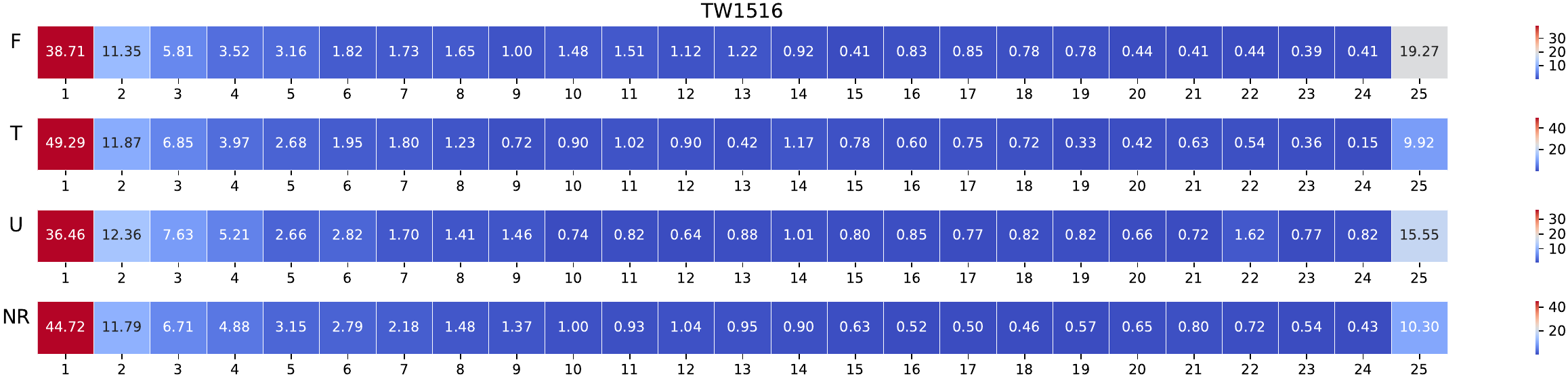}} \\
    \subfloat[PHEME16]{\includegraphics[width=1\textwidth]{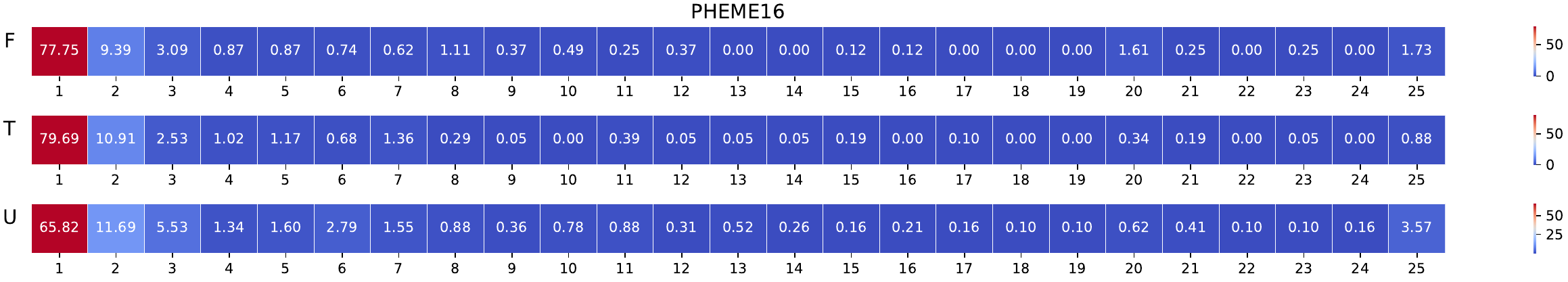}} \\
    \subfloat[RE19]{\includegraphics[width=1\textwidth]{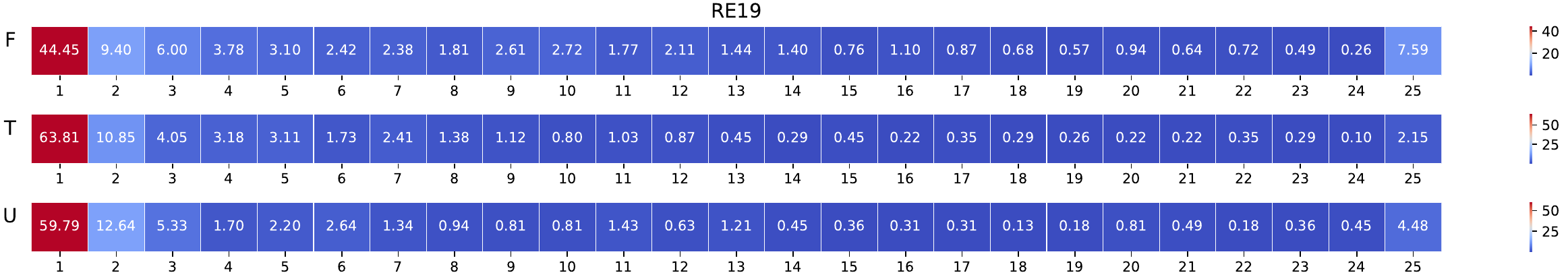}} \\
    \caption{The distribution of comments with time interval (hour). F: False rumor, T: True rumor, U: Unverified rumor, NR: Non-rumor. The last interval (i.e. 25) contains all comments after 24 hours}
    \label{fig:heatmap_fourdatasets}
\end{figure*}

From the distribution of comments with time intervals in Figure \ref{fig:heatmap_fourdatasets}, we can observe that most comments are posted within 24h (more than 80\%). Therefore, we choose to divide the comments within 24 hours and merge all comments after 24 hours into one interval. Table \ref{tab:selectionofhyperpara} presents the performance of different dimensions and time intervals. Considering both the computational complexity of the model and its performance, we ultimately chose a dual dimension of 256 and an interval of 1 hour (i.e. total of 25 intervals, the last interval includes all comments posted after 24h).

After confirming the interval and dual dimension, we conduct experiments on PHEME16, TW1516, RE19 to evaluate the impact of $\lambda$, which controls the magnitude of the auxiliary loss. After comprehensively considering the results in Table \ref{tab:selectionofllambda}, we finally determined that $\lambda$ is 0.5.

\begin{table}[]
\begin{tabular}{lcccc}
\hline
Dimension & ACC   & PRE   & REC   & F1    \\ \hline
32                               & 0.821 & 0.803 & 0.785 & 0.788 \\ 
64                               & 0.802 & 0.778 & 0.774 & 0.773 \\
128                              & 0.804 & 0.782 & 0.780 & 0.778 \\
256                              & 0.827 & 0.808 & 0.800 & 0.800 \\
512                              & 0.838 & 0.828 & 0.802 & 0.808 \\ \hline
Interval                       & ACC   & PRE   & REC   & F1    \\ \hline
1h                              & 0.827 & 0.808 & 0.800 & 0.800 \\
30min                            & 0.825 & 0.801 & 0.784 & 0.787 \\
10min                            & 0.824 & 0.804 & 0.795 & 0.795 \\ \hline
\end{tabular}
\caption{\label{tab:selectionofhyperpara}
Performance of different dual dimensions and time intervals (Conducted on TW1516 dataset). For dual dimensions selection, we set the time interval to 1 hour (i.e. 25 intervals). For interval selection, we set the dual dimension to 256.}
\end{table}

\begin{table}[]
\small
\begin{tabular}{lcccccc}
\hline
                          & \multicolumn{2}{c}{PHEME16} & \multicolumn{2}{c}{TW1516} & \multicolumn{2}{c}{RE19} \\
\multirow{-2}{*}{{\color[HTML]{212121} $\lambda$}} & ACC          & F1           & ACC          & F1          & ACC         & F1         \\ \hline
0                                                  & 0.650        & 0.611        & 0.815        & 0.788       & 0.465       & 0.370      \\
0.2                                                & 0.700        & 0.677        & 0.797        & 0.766       & 0.486       & 0.354      \\
0.5                                                & 0.700        & 0.675        & 0.827        & 0.800       & 0.502       & 0.427      \\
0.8                                                & 0.672        & 0.651        & 0.809        & 0.777       & 0.465       & 0.342      \\
1                                                  & 0.689        & 0.655        & 0.833        & 0.804       & 0.453       & 0.382      \\ \hline
\end{tabular}
\caption{\label{tab:selectionofllambda}
Performance of different $\lambda$ on PHEME16, TW1516, RE19.}
\end{table}

\begin{table}[]
\begin{tabular}{lcccccc}
\hline

\end{tabular}
\end{table}

\section{Prompts for the main task and auxiliary task \label{app:promptexamplestasks}}

\begin{center}
\fcolorbox{black}{gray!10}{
\begin{minipage}{0.45\textwidth}
\footnotesize
\textbf{Template for main task}  \\
\textbf{Task description:} \textit{This is a rumor detection task, you need to judge the source text based on the source text information and corresponding comments (0. false rumour, 1. true rumour,  2. unverified rumour, 3. non-rumours.) }  \\
\textit{The following content is fusion information, including the emotional information and content information of the comments, and the content of the source text information.}  \\
\textit{Please give your answer (0. false rumour, 1. true rumour,  2. unverified rumour, 3. non-rumours.) according to all information.}
\end{minipage}
}
\end{center}

\begin{center}
\fcolorbox{black}{gray!10}{
\begin{minipage}{0.45\textwidth}
\footnotesize
\textbf{Template for auxiliary task}  \\
\textbf{Task description:} \textit{Calculate the sentiment intensity or valence score of the source text, which should be a real number between 0 (extremely negative) and 1 (extremely positive).}  \\
\textit{The following content is fusion information, including the emotional information and content information of the comments, and the content of the source text information.}  \\
\end{minipage}
}
\end{center}

\section{The description of other baselines}
\label{app:baselines}

\begin{itemize}

\item NileTMRG \cite{enayet2017niletmrg} is a linear SVM that uses text features, social features, and comment stance features.

\item Emoratio \cite{ajao2019sentiment} applies sentiment features from news content to detect fake news by calculating affective words.

\item EmoCred \cite{giachanou2019leveraging} leverage sentiment lexicon and sentiment intensity features based on lexicon frequency.

\item AGAT \cite{li2021joint} adopts high-frequency words and stance features to construct a heterogeneous graph for fake news detection.

\item Double-Channel \cite{kim2023detecting} applies a lie detection algorithm to informed rumors and a thread-reply agreement detection algorithm to uninformed rumors.

\item S2MTL-FND \cite{jiang2024makes} is a multitasking learning method using sentiment and stance as auxiliary tasks for misinformation detection.

\end{itemize}


\section{The changes of dual SI difference with time \label{app:sichanges}}

\begin{figure}[htbp]
    \centering
    \subfloat[PHEME18]{\includegraphics[width=0.45\textwidth]{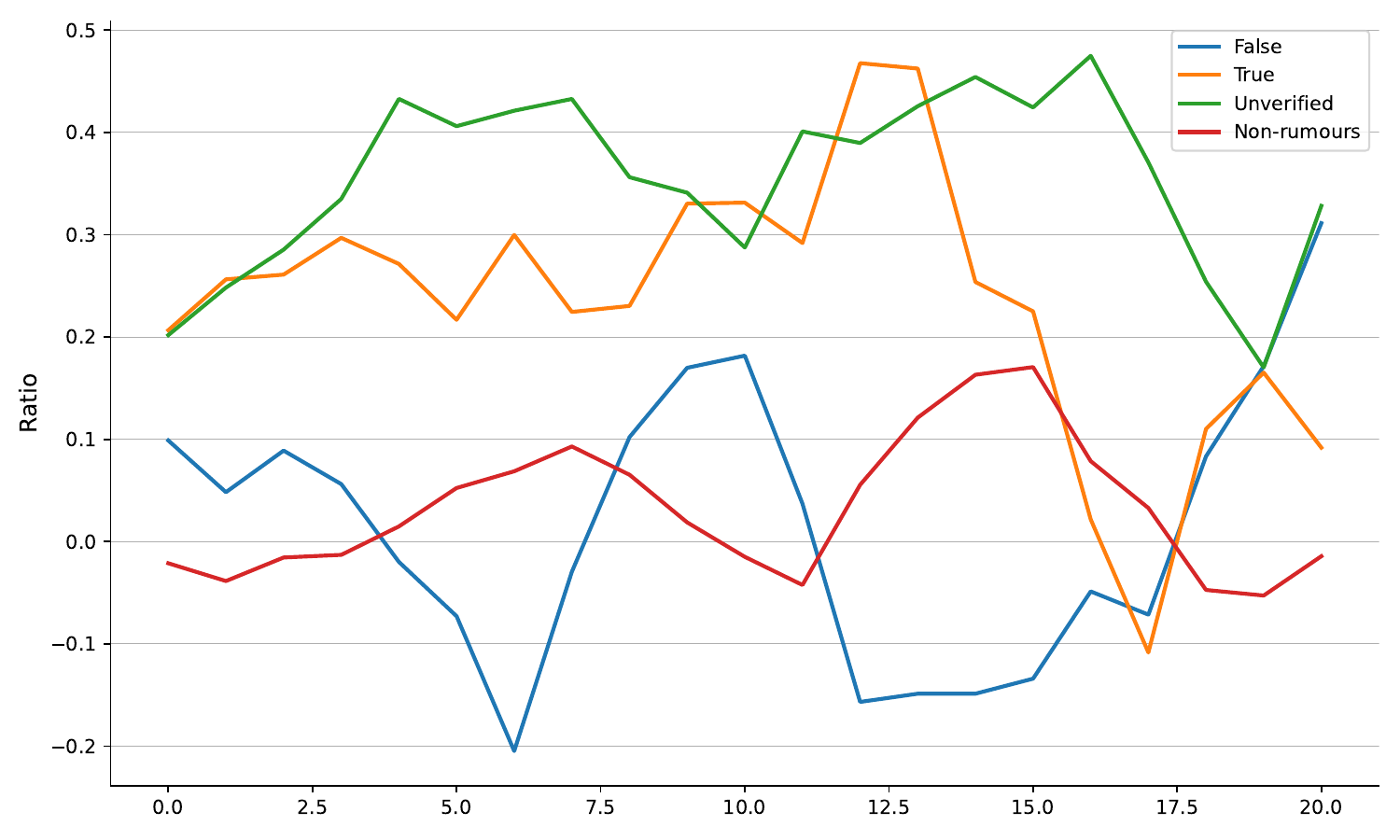}} \\
    \subfloat[TW1516]{\includegraphics[width=0.45\textwidth]{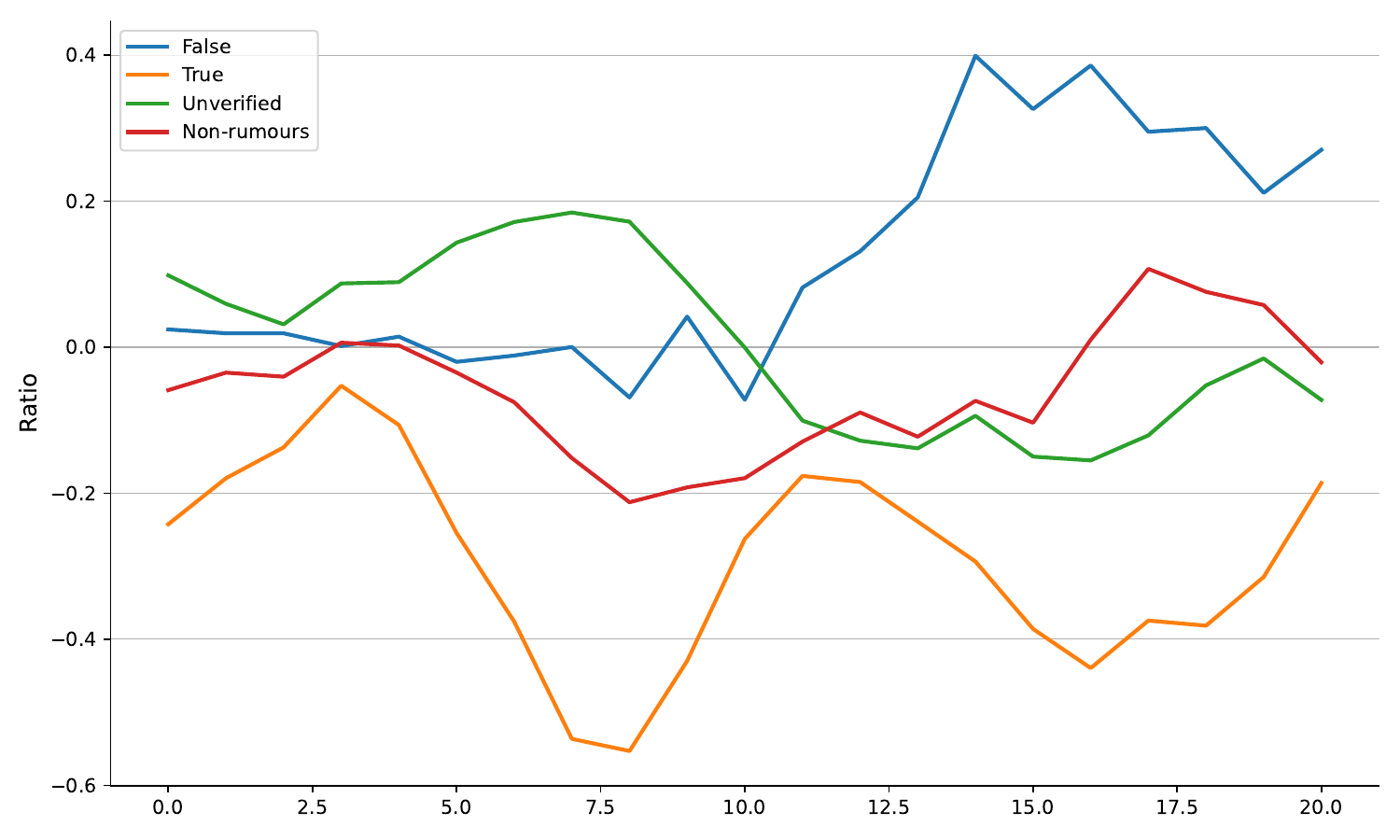}} \\
    \caption{The changes of dual SI difference between two populations ($SI_{dual>0}$ and $SI_{dual<0}$) on the PHEME18 and TW1516 datasets.}
    \label{fig:temporalsentiment}
\end{figure}

Figure \ref{fig:temporalsentiment} presents the dual SI difference with time changes on PHEME18 and TW1516 within 24h (We set the window to 4 for smoothing.). Each value is calculated by ($SI_{dual>0}^{num}$ - $SI_{dual<0}^{num}$) / ($SI_{dual>0}^{num}$ + $SI_{dual<0}^{num}$) (num denotes the number of the specific category). From the figures, we can observe that False rumors (blue) show an overall upward trend in fluctuations, with a larger fluctuation amplitude in PHEME18 and a greater increase in TW1516. True rumors (orange) show an overall downward trend in fluctuations. Unverified rumors (green) show an upward trend in the initial stage, followed by an overall downward trend in the later stage. In contrast, the non-rumor category remains stable, the values close to 0 throughout. There are significant trend differences among different categories within the dataset. Although the trends of different categories show slight variations in different datasets due to different events and debunking times, we can still identify some commonalities from the figures. Overall, these phenomena demonstrate that differences in sentiment intensity can serve an important role in rumor detection.

\section{The distribution of comments with time interval}

\end{document}